\documentclass{article}
\usepackage[preprint,nonatbib]{neurips_2024}
\usepackage[utf8]{inputenc} 
\usepackage[T1]{fontenc}    
\usepackage{hyperref}       
\usepackage{url}            
\usepackage{booktabs}      
\usepackage{amsfonts}       
\usepackage{nicefrac}      
\usepackage{microtype}      
\usepackage{xcolor}         
\usepackage{graphicx}
\usepackage{wrapfig}

\usepackage{amsmath}
\usepackage{enumitem}

\usepackage{comment}
\def\R{\mathbb R}

\title{Equivariant Representation Learning for Augmentation-based Self-Supervised Learning  via Image Reconstruction}

\author{%
  Qin Wang$^{1}$ \quad Kai Krajsek$^{2}$ \quad Hanno Scharr$^{1}$\\
  $^{1}$IAS-8: Data Analytics and Machine Learning,  Forschungszentrum Jülich, Germany \\ 
  $^{2}$Jülich Supercomputing Centre (JSC),  Forschungszentrum Jülich, Germany \\
  \texttt{\{qi.wang, k.krajsek, h.scharr\}@fz-juelich.de}
  } 

\begin{document}

\maketitle

\begin{abstract}
Augmentation-based self-supervised learning methods have shown remarkable success in self-supervised visual representation learning, excelling in learning invariant features but often neglecting equivariant ones. This limitation reduces the generalizability of foundation models, particularly for downstream tasks requiring equivariance. We propose integrating an image reconstruction task as an auxiliary component in augmentation-based self-supervised learning algorithms to facilitate equivariant feature learning without additional parameters. Our method implements a cross-attention mechanism to blend features learned from two augmented views, subsequently reconstructing one of them. This approach is adaptable to various datasets and augmented-pair based learning methods. We evaluate its effectiveness on learning equivariant features through multiple linear regression tasks and downstream applications on both artificial (3DIEBench) and natural (ImageNet) datasets. Results consistently demonstrate significant improvements over standard augmentation-based self-supervised learning methods and state-of-the-art approaches, particularly excelling in scenarios involving combined augmentations. Our method enhances the learning of both invariant and equivariant features, leading to more robust and generalizable visual representations for computer vision tasks.
\end{abstract}

\section{Introduction}
 Popular augmentation-based self-supervised learning methods \cite{chen2020simple, he2020momentum, zbontar2021barlow, vicreg, ibot} have shown remarkable success in their domain, primarily focusing on learning invariant features across different views of the same image. While these approaches have proven to be effective for many tasks, their performance is limited for downstream applications that require equivariant behavior \cite{lee2021improvingtransferabilityrepresentationsaugmentationaware}.

Equivariance in feature learning ensures that a model's learned representations remain consistent under various transformations, including 2D or 3D translations, rotations, scaling, and changes in color or illumination \cite{weiler2021generale2equivariantsteerablecnns}. Mathematically, this property implies that the model's transformation commutes with the transformation acting on both the input and feature spaces. In practical terms, an equivariant model's response to an object in an image remains stable regardless of its position, orientation, or other imaging conditions, potentially leading to better generalization on unseen data.

Recent work, such as SIE (Split Invariant and Equivariant) \cite{garrido2023sie}, has attempted to address the limitations of invariance-focused learning by introducing a split between invariant and equivariant features.
During the pretraining process, SIE \cite{garrido2023sie} uses known transformations (e.g. rotation and colour  jittering) including their parameters to learn a linear mapping between equivariant features from two views. However, this approach faces several challenges:

\begin{itemize}[leftmargin=1em]
    \item It has been tested only on small, artificial datasets (3DIEBench \cite{garrido2023sie}), limiting its proven applicability to real-world scenarios.
    \item It requires prior knowledge of transformations to learn equivariant features, which may not always be available or easily determinable.
    \item It struggles when dealing with images that have undergone unknown transformations.
\end{itemize}

To overcome these limitations, based on recent success in self-supervised learning using reconstruction to learn image features \cite{ gupta2023siamesemaskedautoencoders, mae}, we propose a novel equivariance learning method based on image reconstruction, which leverages a cross-attention mechanism to facilitate the neural network in learning equivariant features. By utilizing image reconstruction, our approach enables the network to better capture the relationships between transformed images, leading to improved learning of equivariant representations. This method can be applied to all natural images without requiring prior knowledge of transformations, such as object motion tracking tasks,  addressing a key limitation of previous work.

Our contributions are as follows:

\begin{itemize}[leftmargin=1em]
    \item We introduce reconstruction as an auxilary task to learn equivariance, addressing the limitations of augmentation-based self-supervised learning.
    \item We demonstrate the effectiveness of our method on both artificial (3DIEBench \cite{garrido2023sie}) and natural (ImageNet \cite{ILSVRC15}) datasets, showing comparable (3DIEBench) and  improved performance (ImageNet) compared to existing baselines.
    \item We provide extensive evaluations on various image transformations, including rotation, color jittering, translation, and scaling, demonstrating the robustness of our learned representations.
\end{itemize}

\section{Method}
\begin{figure}
    \centering
    \includegraphics[width=1\linewidth]{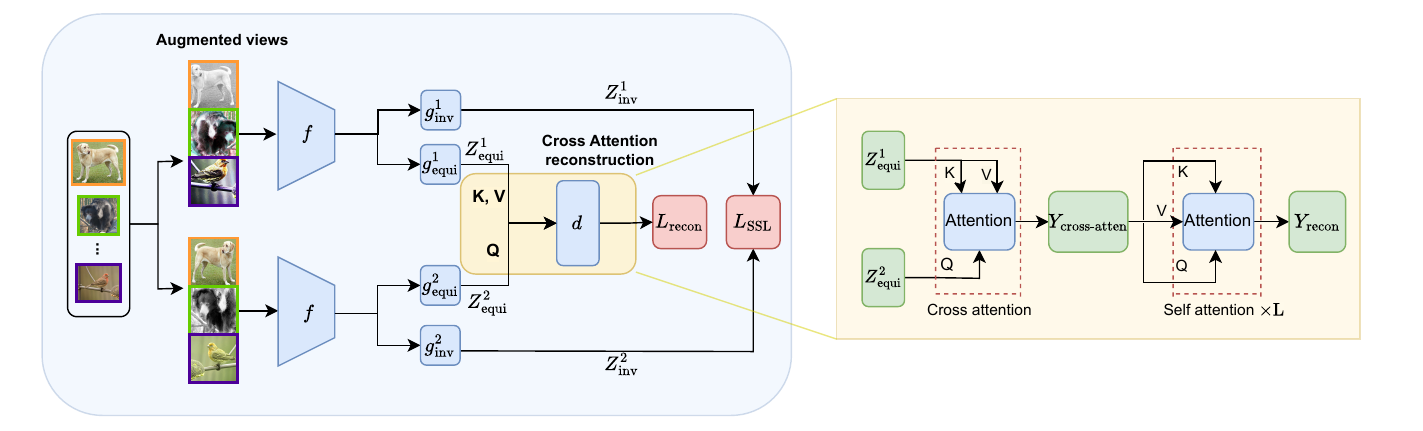}
    \caption{\textbf{Illustration of proposed equivariant reconstruction mechanism.} Cross-attention reconstruction decoder is composed with one cross attention layer, where Key and Value from first view and Query from the second view are mixed for computing the attention matrix. Subsequently, $\text{L} \times$ self-attention layer is added for reconstructing the image.}
    \label{fig:algorithm}
\end{figure}
Our proposed method integrates learning equivariant features with invariant augmentation-based self-supervised learning, as illustrated in Figure \ref{fig:algorithm}. The method builds upon the SIE framework (Split Invariant and Equivariant) \cite{garrido2023sie}. The SIE framework divides the representations extracted from the encoder into two parts: one invariant and the other equivariant. The invariant part uses augmentation-based SSL loss as VICReg \cite{vicreg} to encourage the network to learn invariant features. Meanwhile, the equivariant part first encodes the transformations, enabling the construction of a linear predictor that maps equivariant representations from the first view to the second view. 

In our approach, for the equivariant part, instead of building a linear predictor, the equivariant features are used to compute the reconstruction loss $L_{\text{recon}}$ as the equivariance loss (see details in \ref{sec:cross-atten}). This auxiliary reconstruction task encourages the network to learn robust equivariant features without requiring additional transformation encoding, as is needed in SIE. The final loss is a linear combination of the augmentation-based SSL loss and reconstruction losses, given by $L = \lambda_{\text{SSL}}L_{\text{SSL}} + \lambda_{\text{recon}}L_{\text{recon}}$.

\paragraph{Split Invariant and Equivariant Representations}
\label{sec:split}
The original RGB images, each of size $X \times Y$, in a minibatch of size $N$, denoted as $\text{I} \in \R^{N\times 3 \times X \times Y }$, are augmented to generate two different views, $v_1$ and $v_2$. These views are fed into an encoder, $f$, which shares weights between both inputs. The encoder outputs are then split into two parts: $Y_{\text{inv}}$, which contains invariant information, and $Y_{\text{equi}}$, which contains equivariant information. In our experiments, the output dimension for each view is 512. We split this 512-dimensional vector into two 256-dimensional vectors. These two representations are processed by separate heads, $g_{\text{inv}}$ and $g_{\text{equi}}$, which produce embeddings $Z_{\text{inv}} \in \R^{N \times 192}$ and $Z_{\text{equi}} \in \R^{N \times 192}$, representing the invariant and equivariant features, respectively.

\paragraph{Cross-Attention Reconstruction}
\label{sec:cross-atten}
To facilitate the learning of \textit{equivariant features} from the images, we introduce an auxiliary reconstruction task. The reconstruction is performed using a decoder, $d$, which consists of a \textit{cross-attention layer} followed by $L$ self-attention layers. In the cross-attention layer, the \textit{Key} ($K$) and \textit{Value} ($V$) are derived from $Z_{\text{equi}}^1$ of the first view, $v_1$, while the \textit{Query} ($Q$) is derived from $Z_{\text{equi}}^2$ of the second view, $v_2$. The cross-attention mechanism is defined as:

\begin{equation}
    Y_{\text{cross-atten}} = \text{\textbf{softmax}}((W_Q Z_{\text{equi}}^2)(W_K Z_{\text{equi}}^1)^{\text{T}})(W_V Z_{\text{equi}}^1)
\end{equation}

The output of the cross-attention layer, $Y_{\text{cross-atten}}$, maintains the same dimensionality as the input feature $Z_{\text{equi}}$. This output is then passed through $L$ self-attention layers, yielding the reconstructed images $Y_{\text{recon}}$. These reconstructed images are then used to compute the pixel-wise mean squared error (MSE) as the reconstruction loss $L_{\text{recon}}$, with $v_2$ serving as the target. 

\section{Experimental results}
\textbf{Experiment settings.} We employ the ViT-Small architecture as the base encoder, which is trained with a batch size of $N = 2048$ for 800 epochs. The optimizer used is Adam, with a linearly scaled learning rate of $1 \times 10^{-4}$ and a weight decay of $1 \times 10^{-6}$. The decoder consists of 6 blocks, each with an embedding dimension of 192. Notably, only the first block of the decoder incorporates a cross-attention layer. All experiments adhere to these settings. Under these conditions, pretraining on natural images (ImageNet) takes approximately 23 hours using 16 A100 GPUs.

\textbf{Representation evaluation metrics.}
We follow the same evaluation metrics as SIE \cite{garrido2023sie}, applying a linear classifier on top of the pretrained frozen encoder to predict the transformations. The representations from the two augmented views are fed into a 3-layer MLP, which is trained to regress the transformations between the two views. 

For all transformation predictions, performance is evaluated using the coefficient of determination, $R^2$, which quantifies how well predicted values approximate true values. Specifically, $R^2$ indicates the proportion of variance in the true values that is explained by the predictions, where $R^2 = 1$ represents perfect prediction, and $R^2 = 0$ indicates that the model explains none of the variance.

\label{sec:3die}
\begin{table}[h]
    \centering
    \fontsize{7}{8}\selectfont
    \begin{tabular}{c c c c}
    \hline
    3DIEBench & Classification & Rotation Prediction & Color Prediction \\ \hline
    SIE(rot)    & \textbf{0.820} & \textbf{0.724} & 0.054 \\ 
    SIE(rot+color) & \underline{0.809} & 0.502 & \underline{\textbf{0.980}} \\ 
    Ours  & 0.782 & \underline{0.554} & 0.954 \\ \hline \\
    
    \end{tabular}
    
    \caption{Comparison of different methods on the 3DIEBench \cite{garrido2023sie} dataset. Bold values indicate overall best results, underlined values indicate the better results within direct comparison of Ours and SIE \cite{garrido2023sie} with combined augmentations (rotation and color jittering).}
    \label{tab_3dibench}
\vspace{-16pt}
\end{table}

\textbf{Evaluation on 3DIEBench \cite{garrido2023sie} dataset. }We evaluate our method using the 3DIEBench dataset, which provides transformation parameters. In contrast to the SIE method, which relies on the knowledge of the augmentation transformation parameters, our approach does not require any information about the transformations involved.

From Table \ref{tab_3dibench}, the SIE model with rotation knowledge excels in Classification and Rotation Prediction but performs poorly in Color Prediction without color prior knowledge. Incorporating color augmentation in SIE(rot+color) greatly improves Color Prediction but reduces performance in Rotation Prediction. The Ours model strikes a balance, performing well across all tasks, making it a versatile choice without any knowledge of transformation involved.

\textbf{Evaluation on natural images.}
\label{sec:natural_image}
We augment the image to create two views. SIE needs the augmentation parameters as prior knowledge in the pretraining process. We create augmentation from the first view to the second view, and provide augmentation parameters.

\begin{table}[h]
    \centering
    \fontsize{7}{8}\selectfont
    \begin{tabular}{c c c c c c c }
        \hline
        ImageNet & Rotation & Color  & Blur radius & Translation & Crop prediction & Flip\\ \hline
        SIE(rot)  & \textbf{0.990} & 0.867 & 0.042 & 0.540 & 0.266 & 0.532 \\ 
        SIE(color) & 0.078 & \textbf{0.890} & 0.097 & 0.355 & 0.178 & 0.333 \\ 
        SIE(blur)  & 0.153 & 0.883 & \textbf{0.941} & 0.189 & 0.412 & 0.415 \\ 
        SIE(trans) & 0.213 & 0.885 & 0.023 & \textbf{0.978} & 0.368 & 0.511 \\ 
        SIE(crop)  & 0.273 & 0.819 & 0.018 & 0.450 & \textbf{0.922} & 0.485 \\ 
        SIE(flip)  & 0.155 & 0.798 & 0.056 & 0.312 & 0.266 & \textbf{0.993} \\ 
        \hline
        VICReg\cite{vicreg} & 0.318 $\pm$ 0.005 & 0.804 $\pm$ 0.016& 0.101 $\pm$ 0.023 & 0.333 $\pm$ 0.008 & 0.423 $\pm$ 0.140 & 0.872 $\pm$ 0.070\\
        SIE(all)  & 0.331 $\pm$ 0.007 & 0.899 $\pm$ 0.003 & 0.211 $\pm$ 0.005 & \underline{0.925 $\pm$ 0.002} & 0.835 $\pm$ 0.008 & 0.945 $\pm$ 0.004 \\ 
        SIE(all, single each time)  & 0.435 $\pm$ 0.011 & 0.907 $\pm$ 0.009 & 0.377 $\pm$ 0.004 & 0.922 $\pm$ 0.010 & 0.829 $\pm$ 0.005 & 0.939 $\pm$ 0.007 \\
        Ours       & \underline{0.862 $\pm$ 0.004} & \underline{\textbf{0.921 $\pm$ 0.006}} & \underline{0.823 $\pm$ 0.003} & 0.853 $\pm$ 0.005 & \underline{0.912 $\pm$ 0.002} & \underline{0.952 $\pm$ 0.008} \\ 
        \hline \\
    
    \end{tabular}
    
    \caption{Performance comparison on ImageNet for different prediction tasks. Bold values indicate overall best results, underlined values indicate the better results within direct comparison of Ours and SIE \cite{garrido2023sie}. The results of VICReg are obtained by using a pretrained model with a 3-layer MLP for finetuning, specifically for evaluating equivariance. }
    \label{tab_imagenet}
    \vspace{-16pt}
\end{table}
In Table \ref{tab_imagenet} we see, that SIE models excel when pretrained with specific single transformations, such as rotation, achieving the best results for rotation prediction. However, their performance drops significantly for other transformations. Even with all transformation information (as in SIE(all)), their performance remains lower than ours. Pretraining with randomly selected transformations (as in SIE(all, single each time)) improves results compared to SIE(all) but still falls short of our method.

We further evaluate transfer learning on smaller classification datasets and segmentation tasks, i.e.\ ADE20K\cite{8100027}. The results are attached in \ref{app:trasferlearning}.
\begin{table}[!h]
    \centering
    \fontsize{7}{8}\selectfont
    \begin{tabular}{c c c c c}
    \hline
    Cifar10 & Rotation & Color & Blur Radius & Translation \\ \hline
    Supervised   & 0.214 & 0.229 & 0.437 & 0.386 \\ 
    SIE(all)   & 0.402 & 0.395 & 0.511 & 0.479 \\ 
    %Simclr \\
    Ours       & \textbf{0.815} & \textbf{0.879} & \textbf{0.944} & \textbf{0.878} \\ 
    \hline \\
    \end{tabular}
    \caption{Comparison on partial CIFAR10 data. The bold values indicate the overall best.}
    \label{tab:unseen}
    \vspace{-16pt}
\end{table}

\textbf{Utilisation of unknown transformations for learning equivariant representations}
We evaluate the effectiveness of our method for learning equivariant representations only with knowledge of parts of the augmentation transformations and compare its performance with that of SIE. With the CIFAR10 dataset \cite{Krizhevsky2009LearningML} we denote 80\% of the training data as data subject to unknown transformations and for 20\% the transformations including their parameters are known. Since SIE need to know the transformations, it can only use the 20\% of the data. Therefore, SIE as well as \textit{supervised} are trained exclusively on the remaining 20\% data with known transformations, whereas our method leverages the entire dataset. Both models are evaluated on the validation set. As shown in Table \ref{tab:unseen}, our method significantly outperforms SIE, and thus emphasises its ability to efficiently use data that has been subjected to unknown transformation to generate robust equivariant representations.

\vspace{-6pt}
\section{Conclusions}
\vspace{-6pt}
Our proposed method integrates equivariant representation learning into augmentation-based self-supervised learning through a reconstruction task, demonstrating potential for enhancing the generalization capabilities of invariant augmentation-based self-supervised learning. In this paper, we evaluate our approach on multiple datasets and downstream tasks to measure its impact on the equivariant properties of pretrained networks. Our method matches the performance of SIE \cite{garrido2023sie} on the 3DIEBech dataset and surpasses it on natural image datasets. 

Our experiments are currently limited to using smaller backbones and datasets for testing. In the future, we plan to explore larger network architectures and datasets to further evaluate the effectiveness of our method across a broader range of augmentation-based self-supervised learning techniques. Additionally, we will investigate alternative image reconstruction methods for learning equivariant representations.

{\bf Acknowledgements} The authors gratefully acknowledge the Gauss Centre for Supercomputing e.V. (www.gauss-centre.eu) for funding this project by providing computing time through the John von Neumann Institute for Computing (NIC) on the GCS Supercomputer JUWELS at Jülich Supercomputing Centre (JSC).

\bibliographystyle{IEEEtran}
\bibliography{thebib}

\newpage
\appendix

\section{Appendix / supplemental material}
\subsection{Experiments on CIFAR10}
Similar to the ImageNet dataset discussed in Section \ref{sec:natural_image}, we also perform transformation prediction on smaller datasets, such as CIFAR10. The conclusions drawn from these smaller datasets are consistent with those observed for ImageNet.
\begin{table}[h!]
    \centering
    \fontsize{7}{8}\selectfont
    \begin{tabular}{c c c c c}
    \hline
    Cifar10 & Rot Prediction & Color Prediction & Blur Radius & Trans Prediction \\ \hline
    SIE(rot)  & \textbf{0.989} & 0.887 & 0.836 & 0.911 \\ 
    SIE(color) & 0.813 & \textbf{0.921} & 0.825 & 0.822 \\ 
    SIE(blur)  & 0.814 & 0.833 & \textbf{0.990} & 0.807 \\ 
    SIE(trans) & 0.876 & 0.812 & 0.810 & \textbf{0.987} \\ \hline
    SIE(all)   & \underline{0.845} & 0.864 & 0.889 & 0.886 \\ 
    Ours       & 0.826 & \underline{0.906} & \underline{0.972} & \underline{0.890} \\ 
    \hline \\
    \end{tabular}
    \caption{Comparison of different prediction methods on the CIFAR10 dataset. Bold values indicate overall best results, underlined values indicate the better results within direct comparison of Ours and SIE}
\end{table}
\subsection{Transfer learning on downstream tasks}
\label{app:trasferlearning}
\textbf{Transfer learning on classification tasks.} We follow the standard self-supervised learning evaluation pipeline, where the pretrained network is frozen and only the linear head is fine-tuned on downstream tasks. From Table \ref{tab:performance_classification}, we observe that our method performs well on most classification datasets, with the exception of the Pets dataset, when compared to SIE. In the case of the Aircraft dataset, SIE outperforms other methods due to its rotation prior, which better accommodates the rotation-invariant nature of the images.
\begin{table*}[!h]
    \centering
    \fontsize{7}{8}\selectfont
    \begin{tabular}{cccccccc}
    \hline
         Methods& Cifar10 \cite{Krizhevsky2009LearningML} & Cifar100 \cite{Krizhevsky2009LearningML}& Food101 \cite{food101} & SUN397 \cite{Xiao2010SUNDL} & DTD \cite{dtd} & Pets \cite{pets} & Aircraft \cite{aircraft}  \\
    \hline
         SIE(rot)& 71.56&46.88&55.48&43.11&64.22&81.51&\textbf{50.21 }\\
         SIE(color)& 67.99&48.78&57.19&42.32&60.87&80.27&41.15 \\
         SIE(crop)& 80.84&49.35&59.24&52.38&61.82&84.63&47.35 \\
         \hline
         Supervised& 80.99&50.66&59.32&52.98&62.03&83.59&47.83 \\
         SIE(all)& 79.91$\pm$0.18 & 53.12$\pm$0.05 & 58.42$\pm$0.20 & 56.11$\pm$0.08 & 63.56$\pm$0.11 & \textbf{85.34$\pm$0.19} & 46.88$\pm$0.23 \\
         Ours& \textbf{81.12$\pm$0.11}& \textbf{54.22$\pm$0.10}& \textbf{59.21$\pm$0.14}& \textbf{59.53$\pm$0.13}& \textbf{67.66$\pm$0.12}& 84.32$\pm$0.09& \underline{49.75$\pm$0.22} \\
    \hline \\
    \end{tabular}
    \caption{Transfer learning on classification tasks. Bold values indicate best results within direct comparison of Ours vs. SSL methods, underlined values overall best.}
    \label{tab:performance_classification}
    
\end{table*}

\textbf{Transfer learning on segmentation task.} We use an encoder with HyperNet \cite{7780467} on top of the encoder for the segmentation task. The table below presents the results of transfer learning experiments on the ADE20K dataset, evaluating three different methods: Supervised, SIE(all), and Ours. The metrics used are mean Intersection over Union (mIOU), mean Accuracy (mAcc), and overall Accuracy (aAcc). In this comparison, our method outperforms both the Supervised and SIE(all) approaches across all metrics, achieving the highest mean Intersection over Union (mIOU), mean Accuracy (mAcc), and overall Accuracy (aAcc).
\begin{table}[!h]
    \centering
    \fontsize{7}{8}\selectfont
    \begin{tabular}{c c c c}
    \hline
    ADE20K & mIOU & mAcc & aAcc \\ \hline
    Supervised  & 0.268 & 0.328 & 0.751 \\
    SIE(all)  & 0.292 & 0.356 & 0.774 \\ 
 
    Ours  & \textbf{0.312} & \textbf{0.379} & \textbf{0.802} \\ \hline \\
    \end{tabular}
    \caption{Transfer learning on segmentation tasks. Bold values indicate best results.}
\end{table}

\end{document}